\documentclass[smallextended,twocolumn]{svjour3}

\smartqed

\usepackage{hyperref}
\usepackage{graphicx}
\usepackage{caption}
\usepackage[boxruled,linesnumbered]{algorithm2e}
\usepackage[usenames,dvipsnames]{xcolor}
\usepackage{amsmath}

\usepackage{flushend}

\newcommand{\Probab}[1]{{P}\left({#1}\right)}
\newcommand{\Pcond}[2]{\Probab{{#1}\mid{#2}}}

\renewcommand{\~}[1]{\overline{#1}}

\newcommand{\DAG}{\textsf{DAG}}
\newcommand{\BN}{\textsf{BN}}
\newcommand{\SBCN}{\textsf{SBCN}}
\newcommand{\HIV}{\textsf{HIV}}
\newcommand{\NGS}{\textsf{NGS}}
\newcommand{\PM}{\textsf{PM}}
\newcommand{\MPN}{\textsf{MPN}}
\newcommand{\CMPN}{\textsf{CMPN}}
\newcommand{\DMPN}{\textsf{DMPN}}
\newcommand{\XMPN}{\textsf{XMPN}}
\newcommand{\LL}{\mathcal{L}}
\newcommand{\TP}{\textsf{TP}}
\newcommand{\PR}{\textsf{PR}}

\DeclareMathOperator*{\argmax}{arg\,max}

\newcommand{\cc}{\boldsymbol{c}}

\begin{document}

\title{Modeling cumulative biological phenomena with Suppes-Bayes Causal Networks}

\author{Daniele Ramazzotti \and 
  Alex Graudenzi \and 
  Giulio Caravagna \and 
  Marco Antoniotti
}

\institute{Daniele Ramazzotti \at 
  Department of Pathology, Stanford University, Stanford, CA 94305, USA 
  \email{daniele.ramazzotti@stanford.edu} 
  \and 
  Alex Graudenzi \at 
  Department of Informatics, Systems and Communication, University of Milan-Bicocca, Milan, Italy 
  \and 
  Giulio Caravagna \at 
  School of Informatics, University of Edinburgh, Edinburgh, UK 
  \and 
  Marco Antoniotti \at 
  Department of Informatics, Systems and Communication, University of Milan-Bicocca, Milan, Italy 
}

\date{Received: date / Accepted: date}

\maketitle

\begin{abstract}
Several diseases related to cell proliferation are characterized by the accumulation of somatic DNA changes, with respect to wildtype conditions. Cancer and \HIV{} are two common examples of such diseases, where the mutational load in the cancerous/viral population increases over time. In these cases, selective pressures are often observed along with competition, co-operation and parasitism among distinct cellular clones. Recently, we presented a mathematical framework to model these phenomena, based on a combination of Bayesian inference and Suppes' theory of probabilistic causation, depicted in graphical structures dubbed Suppes-Bayes Causal Networks (\SBCN{}s). \SBCN{}s are generative probabilistic graphical models that recapitulate the potential ordering of accumulation of such DNA changes during the progression of the disease. Such models can be inferred from data by exploiting likelihood-based model-selection strategies with regularization. In this paper we discuss the theoretical foundations of our approach and we investigate in depth the influence on the model-selection task of: $(i)$ the poset based on Suppes' theory and $(ii)$ different regularization strategies. Furthermore, we provide an example of application of our framework to \HIV{} genetic data highlighting the valuable insights provided by the inferred \SBCN{}. 
\end{abstract}

\section{Introduction}\label{sect:introduction}
A number of diseases are characterized by the accumulation of genomic lesions in the DNA of a population of cells. Such lesions are often classified as {\em mutations}, if they involve one or few nucleotides, or {\em chromosomal alterations}, if they involve wider regions of a chromosome. The effect of these lesions, occurring randomly and inherited through cell divisions (i.e., they are {\em somatic}), is that of inducing a phenotypic change in the cells. If the change is advantageous, then the clonal population might enjoy a {\em fitness} advantage over competing clones. In some cases, a natural selection process will tend to select the clones with more advantageous and inheritable traits. This particular picture can be framed in terms of Darwinian evolution as a scenario of survival of the fittest where, however, the prevalence of multiple heterogenous populations is often observed \cite{burrell2013causes}. 

Cancer and \HIV{} are two diseases where the mutational\footnote{From now on, we will use the term mutation to refer to the types of genomic lesions mentioned above.} load in the cancerours/viral population of cells increases over time, and drives phenotypic switches and disease progression. In this paper we specifically focus on these desease, but many biological systems present similar characteristics, see \cite{weinreich2006darwinian,poelwijk2007empirical,lozovsky2009stepwise}. 

The emergence and development of cancer can be characterized as an \emph{evolutionary process} involving a large population of cells, heterogeneous both in their genomes and in their epigenomes. In fact, genetic and epigenetic random alterations commonly occurring in any cell, can occasionally be beneficial to the neoplastic cells and confer to these clones a \emph{functional selective advantage}. During clonal evolution, clones are generally selected for increased proliferation and survival, which may eventually allow the cancer clones to outgrow the competing cells and, in turn, may lead to invasion and metastasis \cite{nowell1976clonal,merlo2006cancer}. By means of such a multi-step stochastic process, cancer cells acquire over time a set of biological capabilities, i.e., \emph{hallmarks} \cite{hanahan2000hallmarks,hanahan2011hallmarks}. However, not all the alterations are involved in this acquisition; as a matter of fact, in solid tumors, we observe an average of $33$ to $66$ genes displaying a somatic mutations \cite{vogelstein2013cancer}. But only some of them are involved in the hallmark acquisition, i.e., \emph{drivers}, while the remaining ones are present in the cancer clones without increasing their \emph{fitness}, i.e., \emph{passengers} \cite{vogelstein2013cancer}. 

The onset of {\sf AIDS} is characterized by the collapse of the immune system after a prolonged asymptomatic period, but its progression's mechanistic basis is still unknown. It was recently hypothesized that the elevated turnover of lymphocytes throughout the asymptomatic period results in the accumulation of deleterious mutations, which impairs immunological function, replicative ability and viability of lymphocytes \cite{galvani2005role,seifert2015framework}. The failure of the modern combination therapies (i.e., highly active antiretroviral therapy) of the disease is mostly due to the capability of the virus to escape from drug pressure by developing drug resistance. This mechanism is determined by \HIV{}'s high rates of replication and mutation. In fact, under fixed drug pressure, these mutations are virtually non-reversible because they confer a strong selective advantage to viral populations \cite{perrin1998hiv,vandamme1999managing}. 

In the last decades huge technological advancements led to the development of {\em Next Generation Sequencing} (\NGS{}) techniques. These allow, in different forms and with different technological characteristics, to read out genomes from single cells or bulk \cite{navin2014cancer,wang2014clonal,gerlinger2012intratumor,gerlinger2014genomic}. Thus, we can use these technologies to quantify the presence of mutations in a sample. With this data at hand, we can therefore investigate the problem of inferring a {\em Progression Model} (\PM{}) that recapitulates the ordering of accumulation of mutations during disease origination and development \cite{caravagna2016picnic}. This problem allows different formulation according to the type of diseases that we are considering, the type of \NGS{} data that we are processing and other factors. We point the reader to \cite{beerenwinkel2015cancer,caravagna2016picnic} for a review on progression models. 

This work is focused on a particular class of mathematical models that are becoming successful to represent such mutational ordering. These are called {\em Suppes-Bayes Causal Networks}\footnote{The first use of these networks appears in \cite{ramazzotti2015capri}, and it's earliest formal definition in \cite{bonchi2015exposing}.} (\SBCN{}, \cite{bonchi2015exposing}), and derive from a more general class of models, Bayesian Networks (\BN{}, \cite{koller2009probabilistic}), that has been successfully exploited to model cancer and \HIV{} progressions \cite{desper1999inferring,beerenwinkel2007conjunctive,gerstung2009quantifying}. \SBCN{}s are probabilistic graphical models that are derived within a statistical framework based on Patrick Suppes' theory of {\em probabilistic causation} \cite{suppes1970probabilistic}. Thus, the main difference between standard Bayesian Networks and \SBCN{}s is the encoding in the model of a set of causal axioms that describe the accumulation process. Both \SBCN{}s and \BN{}s are generative probabilistic models that induces a distribution of observing a particular mutational signature in a sample. But, the distribution induced by a \SBCN{} is also consistent with the causal axioms and, in general, is different from the distribution induced by a standard \BN{} \cite{ramazzotti2015capri}. 

The motivation for adopting a causal framework on top of standard \BN{}s is that, in the particular case of cumulative biological phenomena, \SBCN{}s allow better inferential algorithms and data-analysis pipelines to be developed \cite{loohuis2014inferring,ramazzotti2015capri,caravagna2016picnic}. Extensive studies in the inference of cancer progression have indeed shown that model-selection strategies to extract \SBCN{}s from \NGS{} data achieve better performance than algorithms that infer \BN{}s. In fact, \SBCN{}'s inferential algorithms have higher rate of detection of true positive ordering relations, and higher rate of filtering out false positive ones. In general, these algorithms also show better {\em scalability}, {\em resistance to noise in the data}, and {\em ability to work with datasets with few samples} (see, e.g., \cite{loohuis2014inferring,ramazzotti2015capri}). 

In this paper we give a formal definition of \SBCN{}s, we assess their relevance in modeling cumulative phenomena and investigate the influence of $(a)$ Suppes' poset, and $(b)$ distinct maximum likelihood regularization strategies for model-selection. We do this by carrying out extensive synthetic tests in operational settings that are representative of different possible types of progressions, and data harbouring signals from heterogenous populations. 

\section{Suppes-Bayes Causal Networks}\label{sect:approach}
In \cite{suppes1970probabilistic}, Suppes introduced the notion of \emph{prima facie causation}. A prima facie relation between a cause $u$ and its effect $v$ is verified when the following two conditions are true: $(i)$ \emph{temporal priority} (\TP{}), i.e., any cause happens before its effect and $(ii)$ \emph{probability raising} (\PR{}), i.e., the presence of the cause raises the probability of observing its effect. 

\begin{definition}[Probabilistic causation, \cite{suppes1970probabilistic}] \label{def:praising}
For any two events $u$ and $v$, occurring respectively at times $t_u$ and $t_v$, under the mild assumptions that $0 < \Probab{u}, \Probab{v} < 1$, the event $u$ is called a \emph{prima facie cause} of $v$ if it occurs \emph{before} and \emph{raises the probability} of $u$, i.e., 
\begin{align}
\text{\em \TP{}:} & \quad t_u < t_v \\ 
\text{\em \PR{}:} & \quad \Pcond{v}{u} > \Pcond{v}{\~ u}
\end{align}
\end{definition}

While the notion of prima facie causation has known limitations in the context of the general theories of causality \cite{hitchcock2012probabilistic}, this formulation seems to intuitively characterizes the dynamics of phenomena driven by the {\em monotonic accumulation of events}. In these cases, in fact, a temporal order among the events is implied and, furthermore, the occurrence of an early event {\em positively correlates} to the subsequent occurrence of a later one. Thus, this approach seems appropriate to capture the notion of selective advantage emerging from somatic mutations that accumulate during, e.g., cancer or \HIV{} progression. 

Let us now consider a graphical representation of the aforementioned dynamics in terms of a Bayesian graphical model. 

\begin{definition}[Bayesian network, \cite{koller2009probabilistic}] \label{def:BN} The pair $\mathcal{B}=\langle
G, P \rangle$ is a \emph{Bayesian Network (\BN{})}, where $G$ is a \emph{directed acyclic graph (\DAG{})} $G = (V,E)$ of $V$ nodes and $E$ arcs, and $P$ is a distribution induced over the nodes by the graph. Let $V=\{v_1,\ldots,v_n\}$ be {\em random variables} and the {\em edges/arcs} $E \subseteq V \times V$ encode the conditional dependencies among the variables. Define, for any $v_i \in V$, the {\em parent set} $\pi(v_i) = \{x \mid x\to v_i \in E\}$, then $P$ defines the {\em joint probability distribution} induced by the \BN{} as follow: 
\begin{equation}
\Probab{\mathit{v_1, \ldots, v_n}} = \prod_{v_i \in V} \Probab{v_i \,|\, \pi(v_i)}\, .
\end{equation}
\end{definition}

All in all, a \BN{} is a statistical model which succinctly represents the conditional dependencies among a set of \emph{random variables} $V$ through a \DAG{}. More precisely, a \BN{} represents a factorization of the {\em joint distribution} $\Probab{\mathit{v_1, \ldots, v_n}}$ in terms of marginal (when $\pi(v)=\emptyset$) and conditional probabilities $\Probab{ \cdot \,|\, \cdot}$. 

We now consider a common situation when we deal with data (i.e., observations) obtained at one (or a few) points in time, rather than through a timeline. In this case, we are resticted to work with cross-sectional data, where no explicit information of time is provided. Therefore, we can model the dynamics of cumulative phenomena by means of a specific set of the general BNs where the nodes $V$ represent the accumulating events as \emph{Bernoulli random variables} taking values in \{0, 1\} based on their occurrence: the value of the variable is $1$ if the event is observed and $0$ otherwise. We then define a dataset $D$ of $s$ cross-sectional samples over $n$ Bernoulli random variables as follow. 
 \begin{equation}\label{eq:input}
\bordermatrix{
 & v_1 & v_2 & \cdots & v_n \cr
 s_1 & d_{1,1} & d_{1,2} & \ldots & d_{1,n} \cr
 s_2 & d_{2,1} & d_{2,2} & \ldots & d_{2,n} \cr
 \vdots & \vdots & \vdots & \ddots & \vdots \cr
 s_m & d_{m,1} & d_{m,2} & \ldots & d_{m,n} \cr
 }
= D 
\end{equation}

In order to extend \BN{}s to account for Suppes' theory of probabilistic causation we need to estimate for any variable $v \in V$ its timing $t_v$. 
Because we are dealing with cumulative phenomena and, in the most general case, data that do not harbour any evidente temporal information\footnote{In many cases, the data that we can access are {\em cross-sectional}, meaning that the samples are collected at independent and unknown time-points. For this reason, we have to resort on the simplest possible approach to estimate timings. However, in the case we were provided with explicit observations of time, the temporal priority would be directly and, yet, more efficiently assessable.}, we can use the {\em marginal probability} $P(v)$ as a proxy for $t_v$ (see also the commentary at the end of this section). In cancer and \HIV{}, for instance, this makes sense since mutations are inherited through cells divisions, and thus will fixate in the clonal populations during disease progression, i.e., they are persistent. 

\begin{definition}[Suppes-Bayes Causal Network, \cite{bonchi2015exposing}] \label{def:sbcn} A \BN{} $\mathcal{B}$ is a \SBCN{} if and only if, for any edge $v_i \to v_j \in E$, Suppes' conditions (Definition \ref{def:praising}) hold, that is
 \begin{equation}
 \Probab{v_i} > \Probab{v_j} \quad \text{and} \quad \Pcond{v_j}{v_i} > \Pcond{v_j}{\neg v_i} \,.
\end{equation}
\end{definition}

It should be noted that \SBCN{}s and \BN{}s {\em have the same likelihood function}. Thus, \SBCN{}s {\em do not embed any constraint of the cumulative process in the likelihood computation}, while approaches based on cumulative \BN{}s do \cite{gerstung2009quantifying}. Instead, the structure of the model, $E$, is consistent with the causal model \`{a}-la-Suppes and, of course, this in turn reflects in the induced distribution. Even though this difference seem subtle, this is arguably the most interesting advantage of \SBCN{}s over ad-hoc \BN{}s for cumulative phenomena. 

\paragraph{Model-selection to infer a network from data.}
The structure $G$ of a \BN{} (or of a \SBCN{}) can be inferred from a data matrix $D$, as well as the parameters of the conditional distributions that define $P$. The model-selection task is that of inferring such information from data; in general, we expect different models (i.e., edges) if we infer a \SBCN{} or a \BN{}, as \SBCN{}s encode Suppes' additional constraints. 

The general structural learning, i.e., {\em the model-selection problem}, for \BN{}s is {\sf NP-HARD} \cite{koller2009probabilistic}, hence one needs to resort on approximate strategies. For each \BN{} $\mathcal{B}$ a {\em log-likelihood function} $\LL(D \mid E)$ can be used to search in the space of structures (i.e., the set of edges $E$), together with a regularization function $\mathcal{R}(\cdot)$ that penalizes overly complicated models. The network's structure is then inferred by solving the following optimization problem
\begin{equation}
E_\ast = \argmax_{E} \Big[ \LL(D | E) - \mathcal{R}(E) \Big] \,.
\end{equation}
Moreover, the parameters of the conditional distributions can be computed by {\em maximum-likelihood estimation} for the set of edges $E_\ast$; the overall solution is {\em locally optimal} \cite{koller2009probabilistic}. 

Model-selection for \SBCN{}s works in this very same way, but constrains the search for valid solutions (see, e.g., \cite{ramazzotti2015capri}). In particular, it scans only the subset of edges that are consistent with Definition \ref{def:praising} -- while a \BN{} search will look for the full $V\times V$ space. To filter pairs of edges, Suppes' conditions can be estimated from the data with solutions based, for instance, on bootstrap estimates \cite{ramazzotti2015capri}. The resulting model will satisfy, by construction, the conditions of probabilistic causation. It has been shown that if the underlying phenomenon that produced our data is characterised by an accumulation, then the inference of a \SBCN{}, rather than a \BN{}, leads to {\em much better precision and recall} \cite{loohuis2014inferring,ramazzotti2015capri}. 

We conclude this Section by discussing in details the characteristics of the \SBCN{}s and, in particular, to which extent they are capable of modeling cumulative phenomena. 

\paragraph{Temporal priority.} Suppes' first constraint (``event $u$ is temporally preceding event $v$'', Definition \ref{def:praising}) assumes an underlying {\em temporal (partial) order} $\sqsubseteq_\TP{}$ among the events/ variables of the \SBCN{}, that we need to compute. 

Cross-sectional data, unfortunately, are not provided with an explicit measure of time and hence $\sqsubseteq_\TP{}$ needs to be estimated from data $D$\footnote{We notice that in the case we were provided with explicit observations of time, $\sqsubseteq_\TP{}$ would be directly and, yet, more efficiently assessable.}. The cumulative nature of the phenomenon that we want to model leads to a simple estimation of $\sqsubseteq_\TP{}$: {\em the temporal priority \TP{} is assessed in terms of marginal frequencies \cite{ramazzotti2015capri},}
\begin{equation}
v_j \sqsubseteq_\TP{} v_i \iff
 \Probab{v_i} > \Probab{v_j}
 \, .
\end{equation}
Thus, more frequent events, i.e., $v_i$, are assumed to occur earlier, which is sound when we assume the accumulating events to be irreversible. 

\TP{} is combined with probability raising to complete Suppes' conditions for prima facie (see below). Its contribution is fundamental for model-selection, as we now elaborate. 

First of all, recall that the model-selection problem for \BN{}s is in general {\sf NP-HARD} \cite{koller2009probabilistic}, and that, as a result of the assessment of Suppes' conditions (\TP{} {\em and} \PR{}), we constrain our search space to the networks with a given order. Because of time irreversibility, marginal distributions induce a total ordering $\sqsubseteq_\TP{}$ on the $v_i$, i.e., reflexing $\leq$. Learning \BN{}s given a fixed order $\sqsubseteq$ -- even partial \cite{koller2009probabilistic} -- of the variables bounds the cardinality of the parent set as follows 
\begin{equation}
|\pi(v_x)| \leq |\{ 
v_j \mid v_x \sqsubseteq v_j
\}|\, ,
 \end{equation}
and, in general, it make inference easier than the general case \cite{koller2009probabilistic}. Thus, by constraining Suppes' conditions via $\sqsubseteq_\TP{}$'s total ordering, we drop down the model-selection complexity. It should be noted that, after model-selection, the ordering among the variables that we practically have in the selected arcs set $E$ is in general partial; in the \BN{} literature this is sometimes called {\em poset}. 

\paragraph{Probability raising.} Besides \TP{}, as a second constraint we further require that the arcs are consistent to the condition of \PR{}: this leads to another relation $\sqsubseteq_\PR{}$. Probability raising is equivalent to constraining for {\em positive statistical dependence} \cite{loohuis2014inferring}
\begin{align}
\lefteqn{v_j \sqsubseteq_\PR{} v_i}\\\nonumber
&\quad\quad \iff \Pcond{v_j}{v_i} > \Pcond{v_j}{\~ v_i}\\\nonumber
&\quad\quad \iff \Probab{v_i,v_i} > \Probab{v_i}\Probab{v_j}\, ,
\end{align}
thus we model all and only the positive dependant relations. Definition \ref{def:praising} is thus obtained by selecting those probability raising relations that are consistent with \TP{},
\begin{equation}
 \sqsubseteq_\TP{} \cap \sqsubseteq_\PR{} \, ,
\end{equation}
as the core of Suppes' characterization of causation is relevance \cite{suppes1970probabilistic}. 

If $\sqsubseteq_\TP{}$ reduces the search space of the possible valid structures for the network by setting a specific total order to the nodes, $\sqsubseteq_\PR{}$ instead reduces the search space of the possible valid parameters of the network by requiring that the related conditional probability tables, i.e., $\Probab{\cdot}$, account only for positive statistical dependencies. It should be noted that these constraints affect the structure and the parameters of the model, but the likelihood-function is the same for \BN{}s and \SBCN{}s. 

\paragraph{Network simplification, regularization and spurious causality.} Suppes' criteria are known to be necessary but not sufficient to evaluate general causal claims \cite{suppes1970probabilistic}. Even if we restrict to causal cumulative phenomena, the expressivity of probabilistic causality needs to be taken into account.  

When dealing with small sample sized datasets (i.e., small $m$), many pairs of variables that satisfy Suppes' condition may be {\em spurious causes}, i.e., \emph{false positive}\footnote{An edge is spurious when it satisfies Definition \ref{def:praising}, but it is not actually the true model edge. For instance, for a linear model $u\to v \to w$, transitive edge $u\to w$ is spurious. Small $m$ induces further spurious associations in the data, not necessarily related to particular topological structures.}. \emph{False negatives} should be few, and mostly due to noise in the data. Thus, it follows
\begin{itemize}
\item we expect all the ``statistically relevant'' relations to be also prima facie \cite{ramazzotti2015capri}; 
\item we need to filter out spurious causality instances\footnote{A detailed account of these topics, the particular types of spurious structures and their interpretation for different types of models are available in \cite{loohuis2014inferring,ramazzotti2015capri,ramazzotti2016model}.}, as we know that prima facie overfits .
\end{itemize}

A model-selection strategy which exploits a regularization schema seems thus the best approach to the task. Practically, this strategy will select simpler (i.e., {\em sparse}) models according to a {\em penalized likelihood fit criterion} -- for this reason, it will filter out edges proportionally to how much the regularization is stringent. Also, it will rank spurious association according to a criterion that is consistent with Suppes' intuition of causality, as likelihood relates to statistical (in)dependence. Alternatives based on likelihood itself, i.e., without regularization, do not seem viable to minimize the effect of likelihood's overfit, that happens unless $m\to \infty$ \cite{koller2009probabilistic}. In fact, one must recall that due to statistical noise and sample size, exact statistical (in)dependence between pair of variables is never (or very unlikely) observed. 

\subsection{Modeling heterogeneous populations}
Complex biological processes, e.g., {\em proliferation, nutrition, apoptosis}, are orchestrated by multiple co-operative networks of proteins and molecules. Therefore, different ``mutants'' can evade such control mechanisms in different ways. Mutations happen as a random process that is unrelated to the relative fitness advantage that they confer to a cell. As such, different cells will deviate from wildtype by exhibiting different mutational signatures during disease progression. This has an implication for many cumulative disease that arise from populations that are {\em heterogeneous}, both at the level of the single patient (intra-patient heterogeneity) or in the population of patients (inter-patient heterogeneity). Heterogeneity introduces significant challenges in designing effective treatment strategies, and major efforts are ongoing at deciphering its extent for many diseases such as cancer and \HIV \cite{beerenwinkel2007conjunctive,ramazzotti2015capri,caravagna2016picnic}. 

We now introduce a class of mathematical models that are suitable at modeling heterogenous progressions. These models are derived by augmenting \BN{}s with logical formulas, and are called \emph{Monotonic Progression Networks} (\MPN{}s) \cite{farahani2013learning,korsunsky2014inference}. \MPN{}s represent the progression of events that accumulate {\em monotonically}\footnote{The events accumulate over time and when later events occur earlier events are observed as well.} over time, where the conditions for any event to happen is described by a probabilistic version of the canonical boolean operators, i.e., conjunction ($\land$), inclusive disjunction ($\lor$), and exclusive disjunction ($\oplus$). 

Following \cite{farahani2013learning,korsunsky2014inference}, we define one type of \MPN{}s for each boolean operator: the {\em conjunctive} (\CMPN{}), the {\em disjunctive semi-monotonic} (\DMPN{}), and the {\em exclusive disjunction} (\XMPN{}). The operator associated with each network type refers to the logical relation among the parents that eventually lead to the common effect to occur. 

\begin{definition}[Monotonic Progression Networks, \cite{farahani2013learning,korsunsky2014inference}] \label{def:MPN} \MPN{}s are \BN{}s that, for $\theta, \epsilon \in [0,1]$ and $\theta \gg \epsilon$, satisfy the conditions shown in Table~\ref{table:xPN-defs} for each $v\in V$.
\begin{table*}
\begin{align}
\CMPN{}: 
&& \Pcond{v}{\sum \pi(v) = |\pi(v)|} = \theta &&
\Pcond{v}{\sum \pi(v) < |\pi(v)|} \le \epsilon\, , \\
\DMPN{}: &&
\Pcond{v}{\sum \pi(v) > 0} = \theta &&
\Pcond{v}{\sum \pi(v) = 0} \le \epsilon\, , \\
\XMPN{}: &&
\Pcond{v}{\sum \pi(v) = 1} = \theta &&
\Pcond{v}{\sum \pi(v) \neq 1} \le \epsilon\, .
\end{align}
\caption{Definitions for \CMPN{}, \DMPN{}, and \XMPN{}.}
\label{table:xPN-defs}
\end{table*}
\end{definition}
Here, $\theta$ represents the conditional probability of any ``effect'' to follow its preceding ``cause'' and $\epsilon$ models the probability of any noisy observation -- that is the observation of a sample where the effects are observed without their causes. Note that the above inequalities define, for each type of \MPN{}, specific constraints to the induced distributions. These are sometimes termed, according to the probabilistic logical relations, \emph{noisy-AND}, \emph{noisy-OR} and \emph{noisy-XOR} networks \cite{pearl2014probabilistic,korsunsky2014inference}. 

\paragraph{Model-selection with heterogeneous populations.}
When dealing with heterogeneous populations, the task of model selection, and, more in general, any statistical analysis, is non-trivial. One of the main reason for this state of affairs is the emergence of statistical paradoxes such as Simpson's paradox \cite{yule1903notes,simpson1951interpretation}. This phonomenon refers to the fact that sometimes, associations among dichotomous variables, which are similar within subgroups of a population, e.g., females and males, change their statistical trend if the individuals of the subgroups are pooled together. Let us know recall a famous example to this regard. The admissions of the University of Berkeley for the fall of $1973$ showed that men applying were much more likely than women to be admitted with a difference that was unlikely to be due to chance. But, when looking at the individual departments separately, it emerged that $6$ out of $85$ were indead biased in favor of women, while only $4$ presented a slighly bias against them. The reason for this inconsistency was due to the fact that women tended to apply to competitive departments which had low rates of admissions while men tended to apply to less-competitive departments with high rates of admissions, leading to an apparent bias toward them in the overall population \cite{bickel1975sex}. 

Similar situations may arise in cancer when different populations of cancer samples are mixed. As an example, let us consider an hypothetical progression leading to the alteration of gene $e$. Let us now assume that the alterations of this gene may be due to the previous alterations of either gene $c_1$ or gene $c_2$ exclusively. If this was the case, then we would expect a significant pattern of selective advantage from any of its causes to $e$ if we were able to stratify the patients accordingly to either alteration $c_1$ or $c_2$, but we may lose these associations when looking at all the patients together. 

In \cite{ramazzotti2015capri}, the notion of \emph{progression pattern} is introduced to describe this situation, defined as a boolean relation among all the genes, members of the parent set of any node as the ones defined by \MPN{}s. To this extent, the authors extend Suppes' definition of prima facie causality in order to account for such patterns rather than for relations among atomic events as for Definition \ref{def:praising}. Also, they claim that general \MPN{}s can be learned in polynomial time provided that the dataset given as input is \emph{lifted} \cite{ramazzotti2015capri} with a Bernoulli variable per causal relation representing the logical formula involving any parent set. 

Following \cite{ramazzotti2015capri,ramazzotti2016model}, we now consider any formula in {\em conjunctive normal form} (CNF)
\[
\varphi=\cc_1 \wedge \ldots \wedge \cc_n,
\]
where each $\cc_i$ is a {\em disjunctive clause} $\cc_i=c_{i,1} \vee \ldots \vee c_{i,k}$ over a set of literals and each literal represents an event (a Boolean variable) or its negation. By following analogous arguments as the ones used before, we can extend Definition \ref{def:praising} as follows. 

\begin{definition}[CNF probabilistic causation, \cite{ramazzotti2015capri,ramazzotti2016model}] \label{def:praising-general}
For any CNF formula $\varphi$ and $e$, occurring respectively at times $t_\varphi$ and $t_e$, under the mild assumptions that $0 < \Probab{\varphi}, \Probab{e} < 1$, $\varphi$ is a \emph{prima facie cause} of $e$ if 
\begin{equation}
t_\varphi < t_e \quad \text{and} \quad \Pcond{e}{\varphi} > \Pcond{e}{\~ \varphi} \,.
\end{equation}
\end{definition}

Given these premises, we can now define the Extended Suppes-Bayes Causal Networks, an extension of \SBCN{}s which allows to model heterogeneity as defined probabilistically by \MPN{}s. 

\begin{definition}[Extended Suppes-Bayes Causal Network] \label{def:esbcn} A \BN{} $\mathcal{B}$ is an Extended \SBCN{} if and only if, for any edge $\varphi_i \to v_j \in E$, Suppes' generalized conditions (Definition \ref{def:praising-general}) hold, that is
 \begin{equation}
 \Probab{\varphi_i} > \Probab{v_j} \quad \text{and} \quad \Pcond{v_j}{v_\varphi} > \Pcond{v_j}{\neg v_\varphi} \,.
\end{equation}
\end{definition}

\section{Evaluation on simulated data}\label{sect:results}
We now evaluate the performance of the inference of Suppes-Bayes Causal Network on simulated data, with specific attention on the impact of the constraints based on Suppes' probabilistic causation on the overall performance. All the simulations are performed with the following settings. 

We consider $6$ different topological structures: the first two where any node has at the most one predecessor, i.e., $(i)$ trees, $(ii)$ forests, and the others where we set a limit of $3$ predecessors and, hence, we consider $(iii)$ directed acyclic graphs with a single source and conjunctive parents, $(iv)$ directed acyclic graphs with multiple sources and conjunctive parents, $(v)$ directed acyclic graphs with a single source and disjunctive parents, $(vi)$ directed acyclic graphs with multiple sources and disjunctive parents. For each of these configurations, we generate $100$ random structures. 

Moreover, we consider $4$ different sample sizes ($50$, $100$, $150$ and $200$ samples) and $9$ noise levels (i.e., probability of a random entry for the observation of any node in a sample) from $0\%$ to $20\%$ with step $2.5\%$. Furthermore, we repeat the above settings for networks of $10$ and $15$ nodes. Any configuration is then sampled $10$ times independently, for a total of more than $4$ million distinct simulated datasets. 

Finally, the inference of the structure of the \SBCN{} is performed using the algorithm proposed in \cite{ramazzotti2015capri} and the performance is assessed in terms of: $accuracy = \frac{(TP + TN)}{(TP + TN + FP + FN)}$, $sensitivity = \frac{TP}{(TP + FN)}$ and $specificity = \frac{TN}{(FP + TN)}$ with $TP$ and $FP$ being the true and false positive (we define as positive any arc that is present in the network) and $TN$ and $FN$ being the true and false negative (we define negative any arc that is not present in the network). All these measures are values in $[0,1]$ with results close to $1$ indicators of good performance. 

\begin{figure*}[!t]
\center
\includegraphics[width=0.99\textwidth]{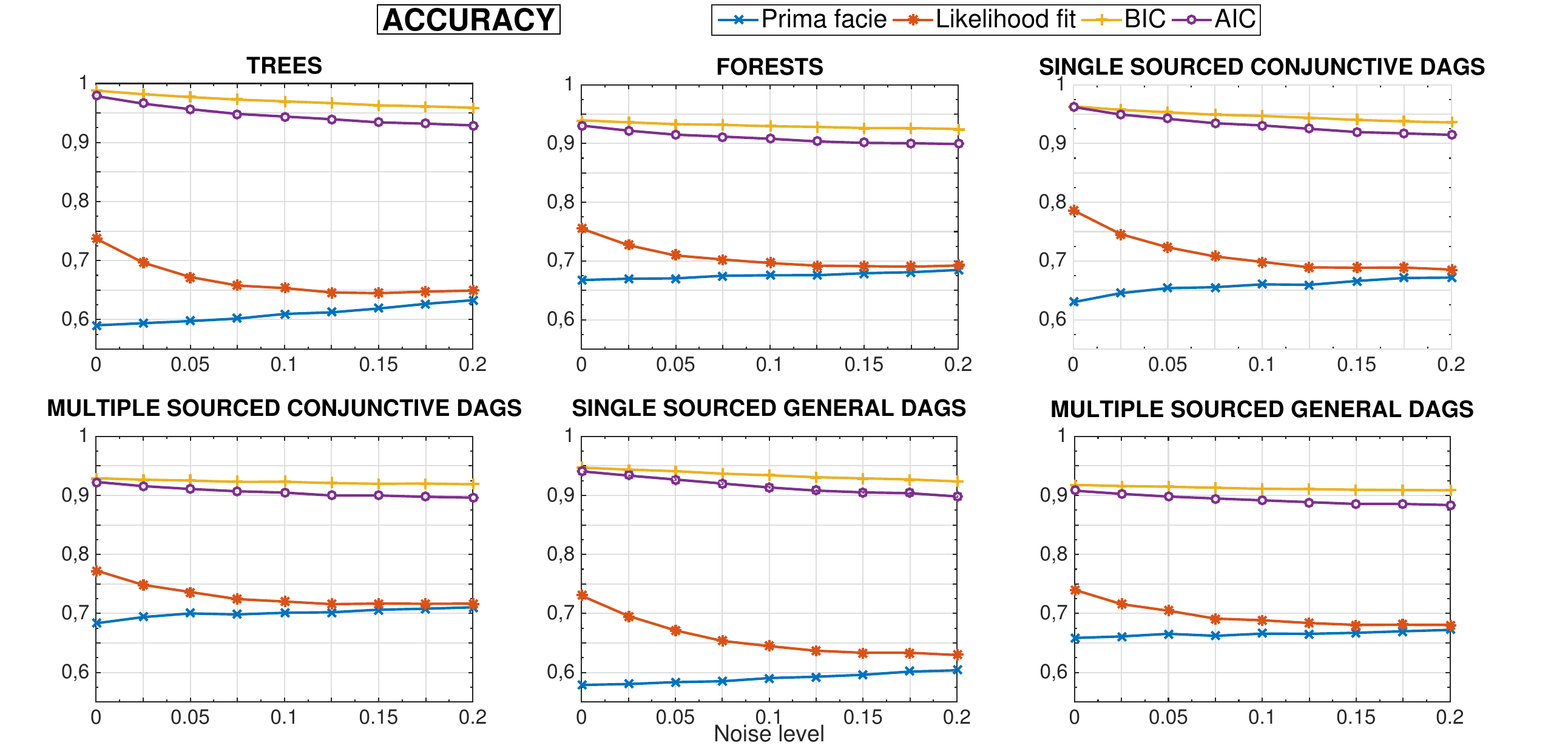}
\caption{Performance of the inference on simulated datasets of $100$ samples and networks of $15$ nodes in terms of accurancy for the $6$ considered topological structures. The $y$ axis refers to the performance while the $x$ axis represent the different noise levels.}
\label{fig:performance_accurancy}
\end{figure*}

\begin{figure*}[!t]
\center
\includegraphics[width=0.99\textwidth]{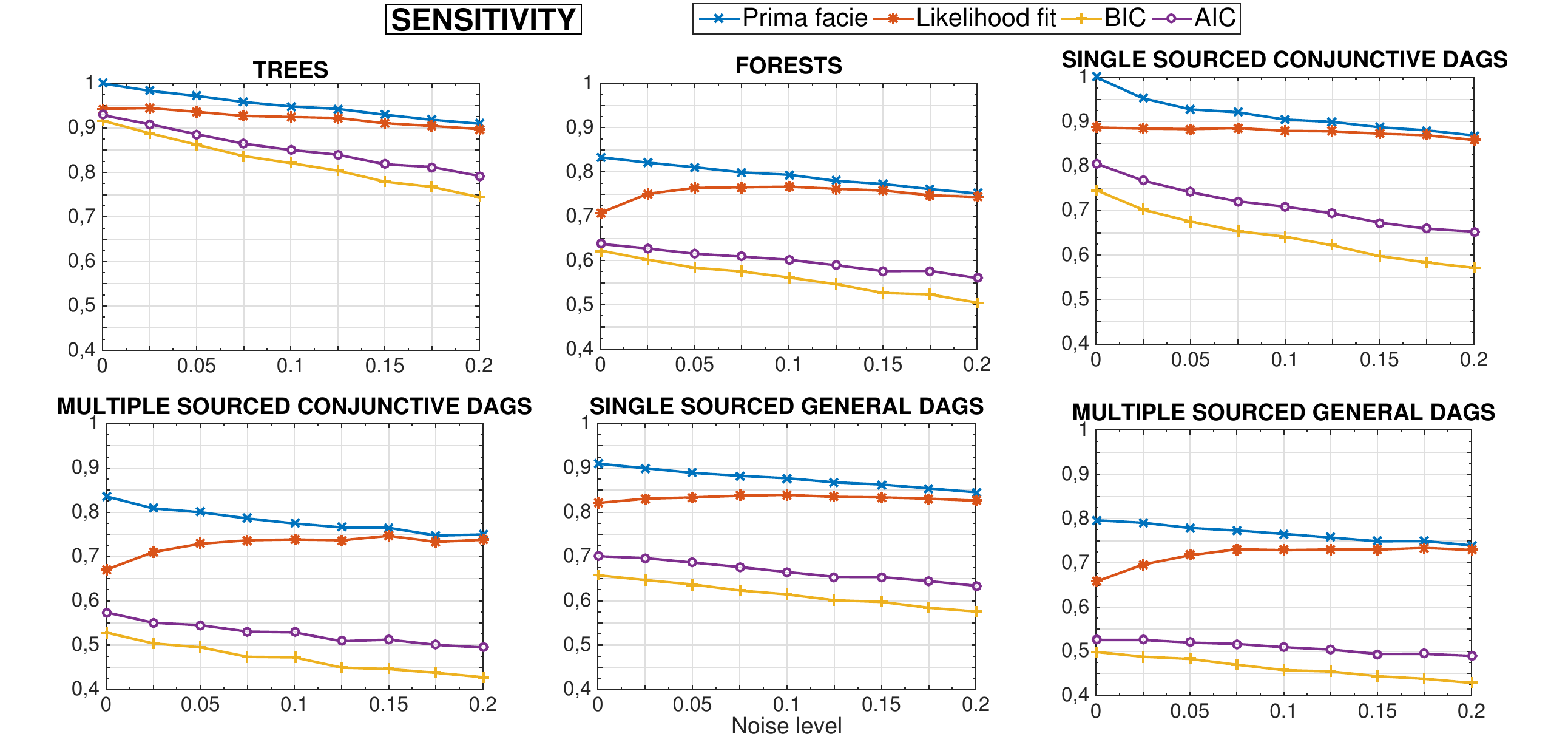}
\caption{Performance of the inference on simulated datasets of $100$ samples and networks of $15$ nodes in terms of sensitivity for the $6$ considered topological structures. The $y$ axis refers to the performance while the $x$ axis represent the different noise levels.}
\label{fig:performance_sensitivity}
\end{figure*}

\begin{figure*}[!t]
\center
\includegraphics[width=0.99\textwidth]{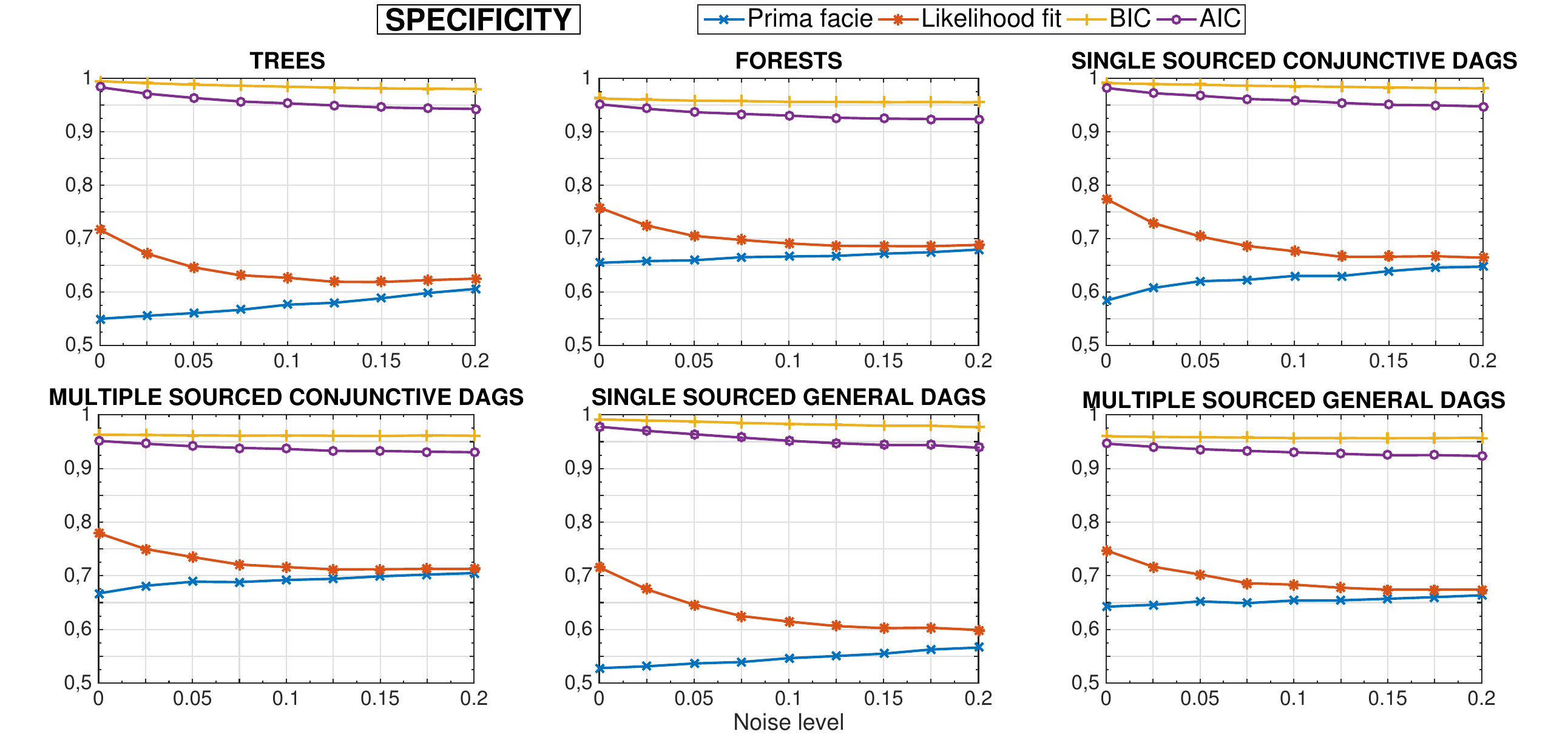}
\caption{Performance of the inference on simulated datasets of $100$ samples and networks of $15$ nodes in terms of specificity for the $6$ considered topological structures. The $y$ axis refers to the performance while the $x$ axis represent the different noise levels.}
\label{fig:performance_specificity}
\end{figure*}

In Figures \ref{fig:performance_accurancy}, \ref{fig:performance_sensitivity} and \ref{fig:performance_specificity} we show the performance of the inference on simulated datasets of $100$ samples and networks of $15$ nodes in terms of accurancy, sensitivity and specificity for different settings which we discuss in details in the next Paragraphs. 

\paragraph{Suppes' prima facie conditions are necessary but not sufficient}
We first discuss the performance by applying \emph{only} the prima facie criteria and we evaluate the obtained prima facie network in terms of accurancy, sensitivity and specificity on simulated datasets of $100$ samples and networks of $15$ nodes (see Figures \ref{fig:performance_accurancy}, \ref{fig:performance_sensitivity} and \ref{fig:performance_specificity}). As expected, the sensitivity is much higher than the specificity implying the significant impact of false positives rather than false negatives for the networks of the prima facie arcs. This result is indeed expected being Suppes' criteria mostly capable of removing some of the arcs which do not represent valid causal relations rather than asses the exact set of valid arcs. Interestingly, the false negatives are still limited even when we consider DMPN, i.e., when we do not have guarantees for the algoritm of \cite{ramazzotti2015capri} to converge. The same simulations with different sample sizes ($50$, $150$ and $200$ samples) and on networks of $10$ nodes present a similar trend (results not shown here). 

\paragraph{The likelihood score overfits the data}
In Figures \ref{fig:performance_accurancy}, \ref{fig:performance_sensitivity} and \ref{fig:performance_specificity} we also show the performance of the inference by likelihood fit (without any regularizator) on the prima facie network in terms of accurancy, sensitivity and specificity on simulated datasets of $100$ samples and networks of $15$ nodes. Once again, in general, sensitivity is much higher than specificity implying also in this case a significant impact of false positives rather than false negatives for the inferred networks. These results make explicit the need for a regularization heuristic when dealing with real (not infinite) sample sized datasets as discussed in the next paragraph. Another interesting consideration comes from the observation that the prima facie networks and the networks inferred via likelihood fit without regularization seem to converge to the same performance as the noise level increases. This is due to the fact that, in general, the prima facie constraints are very conservative in the sense that false positives are admitted as long as false negatives are limited. When the noise level increases, the positive dependencies among nodes are generally reduced and, hence, less arcs pass the prima facie cut for positive dependency. Also in this case, the same simulations with different sample sizes ($50$, $150$ and $200$ samples) and on networks of $10$ nodes present a similar trend (results not shown here). 

\paragraph{Model-selection with different regularization strategies}
We now investigate the role of different regularizations on the performance. In particular, we consider two commonly used regularizations: $(i)$ the \emph{Bayesian information criterion} (BIC) \cite{schwarz1978estimating} and $(ii)$ the \emph{Akaike information criterion} (AIC) \cite{akaike1998information}. 

Although BIC and AIC are both scores based on maximum likelihood estimation and a penalization term to reduce overfitting, yet with distinct approaches, they produce significantly different behaviors. More specifically, BIC assumes the existence of one \emph{true} statistical model which is generating the data while AIC aims at finding the best approximating model to the unknown data generating process. As such, BIC may likely underfit, whereas, conversely, AIC might overfit\footnote{Thus, BIC tends to make a trade-off between the likelihood and model complexity with the aim of inferring the statistical model which generates the data. This makes it useful when the purpose is to detect the best model describing the data. Instead, asymptotically, minimizing AIC is equivalent to minimizing the cross validation value \cite{stone1977asymptotic}. It is this property that makes the AIC score useful in model selection when the purpose is prediction. Overall, the choise of the regularizator tunes the level of sparsity of the retrieved \SBCN{} and, yet, the confidence of the inferred arcs.}. 

The performance on simulated datasets are shown in Figure \ref{fig:performance_accurancy}, \ref{fig:performance_sensitivity}, \ref{fig:performance_specificity}. In general, the performance is improved in all the settings with both regularizators, as they succeed in shrinking toward sparse networks. 

Furthermore, we observe that the performance obtained by \SBCN{}s is still good even when we consider simulated data generated by \DMPN{}. Although in this case we do not have any guarantee of convergence, in practice the algorithm seems efficient in approximating the generative model. In conclusion, without any further input, \SBCN{}s can model \CMPN{}s and, yet, depict the more significant arcs of \DMPN{}s. To infer \XMPN{}, the dataset needs to be lifted \cite{ramazzotti2015capri}. 

The same simulations with different sample sizes ($50$, $150$ and $200$ samples) and on networks of $10$ nodes present a similar trend (results not shown here). 

\section{Application to HIV genetic data}\label{sect:results_real_data}
We now present an example of application of our framework on \HIV{} genomic data. In particular we study drug resistance in patients under antiretroviral therapy and we select a set of $7$ amino acid alterations in the \HIV{} genome to be depicted in the resulting graphical model, namely $K20R$, $M36I$, $M46I$, $I54V$, $A71V$, $V82A$, $I84V$, where, as an example, the genomic event $K20R$ describes a mutation from lysine ($K$) to arginine ($R$) at position 20 of the $HIV$ protease. 

\begin{figure*}[!t]
\center
\includegraphics[width=0.99\textwidth]{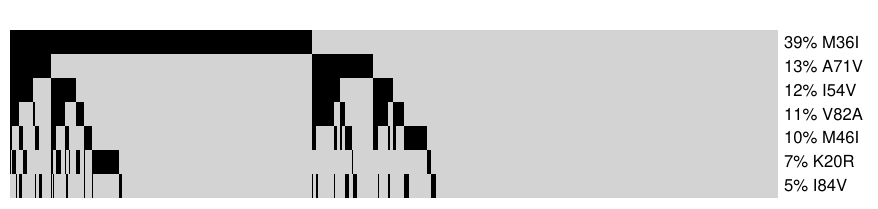}
\includegraphics[width=0.99\textwidth]{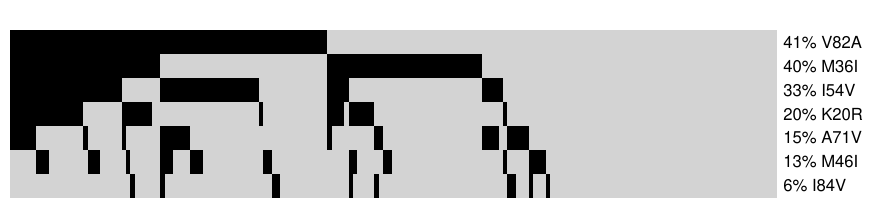}
\caption{Mutations detected in the genome for $179$ \HIV{} patients under ritonavir (top), and $1035$ under indinavir (bottom). Each black rectangle denotes the presence of a mutation in the gene annotated to the right of the plot; percentages correspond to marginal probabilities.}
\label{fig:oncoprint}
\end{figure*}

In this study, we consider datasets from the Stanford HIV Drug Resistance Database (see, \cite{rhee2003human}) for $2$ protease inhibitors, ritonavir ($RTV$) and indinavir ($IDV$). The first dataset consists of $179$ samples (see Figure \ref{fig:oncoprint}) and the second of $1035$ samples (see Figure \ref{fig:oncoprint}). 

\begin{figure*}[!t]
\center
\includegraphics[width=0.99\textwidth]{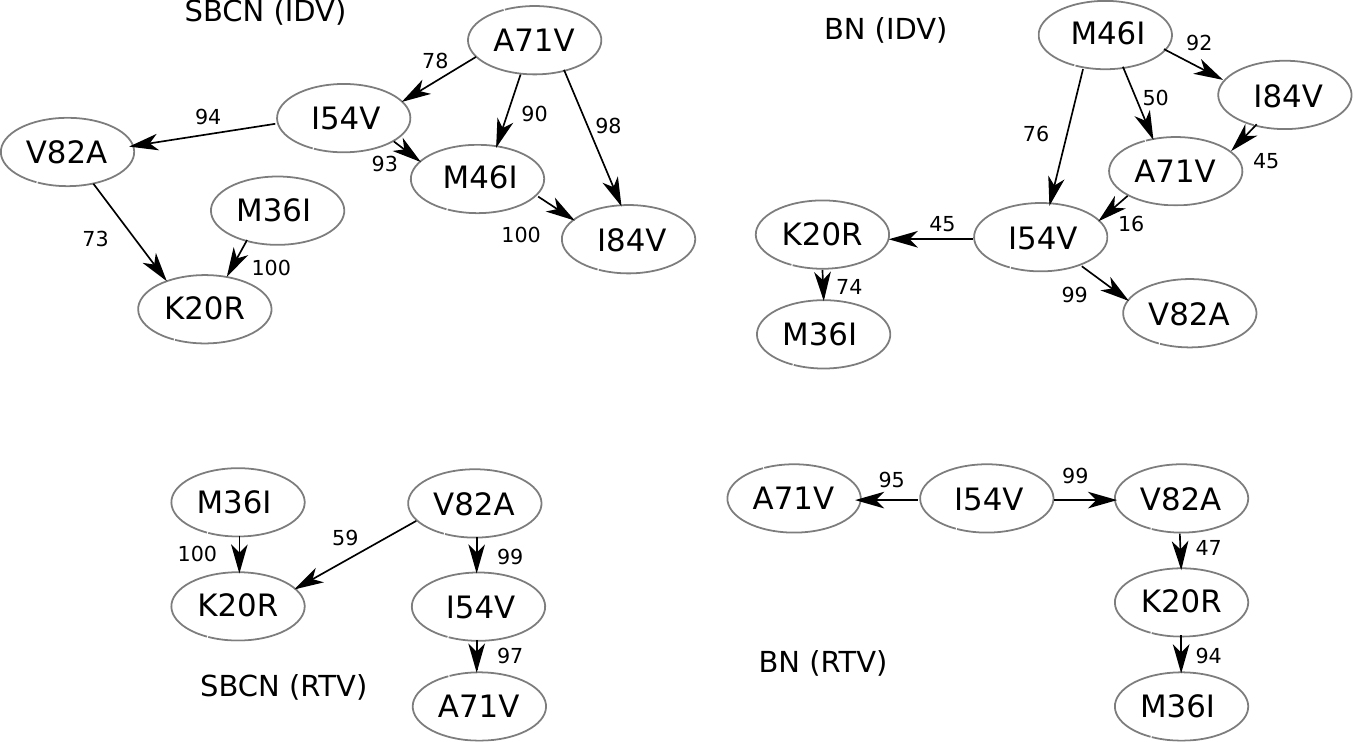}
\caption{\HIV{} progression of patients under ritonavir or indinavir, (Figure \ref{fig:oncoprint}) described as a Bayesian Network or as a Suppes-Bayes Causal Network. Edges are annotated with non parametric bootstrap scores.}
\label{fig:HIV_reconstruction}
\end{figure*}

We then infer a model on these datasets by both Bayesian Network and Suppes-Bayes Causal Network. We show the results in Figures \ref{fig:HIV_reconstruction} where each node represent a mutation and the scores on the arcs measure the confidence in the found relation by non-parametric bootstrap. 

In this case, it is interesting to observe that the set of dependency relations (i.e., any pair of nodes connected by an arc, without considering its direction) depicted both by \SBCN{}s and \BN{}s are very similar, with the main different being in the direction of some connection. This difference is expected and can be attributed to the constrain of temporal priority adopted in the \SBCN{}s. Furthermore, we also observe that most of the found relations in the \SBCN{} are more confidence (i.e., higher bootstrap score) than the one depicted in the related \BN{}, leading us to observe a higher statistical confidence in the models inferred by \SBCN{}s. 

\section{Conclusions}\label{sect:conclusions}
In this work we investigated the properties of a constrained version of Bayesian network, named \SBCN{}, which is particularly sound in modeling the dynamics of system driven by the monotonic accumulation of events, thanks to encoded poset based on Suppes' theory of probabilistic causation. In particular, we showed how \SBCN{}s can, in general, describe different types of MPN, which makes them capable of characterizing a broad range of cumulative phenomena not limited to cancer evolution and \HIV{} drug resistance. 

Besides, we investigated the influence of Suppes' poset on the inference performance with cross-sectional synthetic datasets. In particular, we showed that Suppes' constraints are effective in defining a partially order set accounting for accumulating events, with very few false negatives, yet many false positives. 
To overcome this limitation, we explored the role of two maximum likelihood regularization parameters, i.e., BIC and AIC, the former being more suitable to test previously conjectured hypotheses and the latter to predict novel hypotheses. 

Finally, we showed on a dataset of HIV genomic data how \SBCN{} can be effectively adopted to model cumulative phenomena, with results presenting a higher statistical significance compared to standard \BN{}s. 

\begin{acknowledgements}
This work has been partially supported by grants from the SysBioNet project, a MIUR initiative for the Italian Roadmap of European Strategy Forum on Research Infrastructures (ESFRI). 
\end{acknowledgements}

\label{sect:bib}
\bibliographystyle{spmpsci}
\bibliography{biblio}

\end{document}